\documentclass[runningheads]{llncs}

\usepackage{accv}

\usepackage{graphicx}
\usepackage{booktabs}
\usepackage{amsmath,amssymb}
\usepackage[table]{xcolor}
\usepackage{caption}
\usepackage{enumitem}
\usepackage{url}
\usepackage[accsupp]{axessibility}

\usepackage{tabularx}
\usepackage{array}
\usepackage{siunitx}
\usepackage{float}
\usepackage{placeins}
\usepackage{xspace}
\usepackage[most]{tcolorbox}

\tcbuselibrary{raster}
\usepackage{accvabbrv}

\usepackage[pagebackref,breaklinks,colorlinks,citecolor=accvblue]{hyperref}

\usepackage{orcidlink}

\setlist{nosep,leftmargin=1.2em}
\newcommand{\ci}[2]{{\tiny\textcolor{black!55}{[#1--#2]}}}
\newcommand{\scoreci}[3]{#1\,\ci{#2}{#3}}
\newcommand{\bestscoreci}[3]{\textbf{#1}\,\ci{#2}{#3}}
\newcommand{\secondscoreci}[3]{\underline{#1}\,\ci{#2}{#3}}

\newcommand{\rgain}[1]{\textcolor{black!55}{#1}}

\definecolor{topgreen}{RGB}{0,110,60}

\newcommand{\dataset}{Front2Back-ReID\xspace}

\newcommand{\modelmark}[2]{\raisebox{-0.18\height}{#1}\,\mbox{#2}}
\newcommand{\QwenModel}[1]{\modelmark{\includegraphics[height=0.34cm]{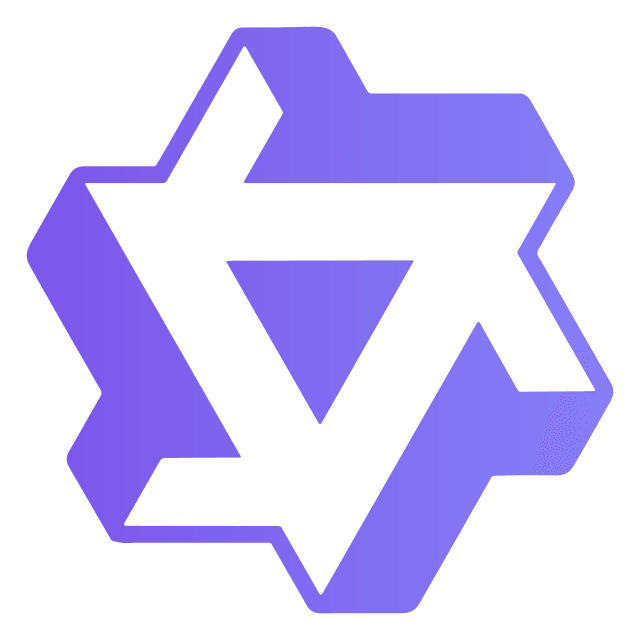}}{#1}}
\newcommand{\LlamaModel}[1]{\modelmark{\includegraphics[height=0.34cm]{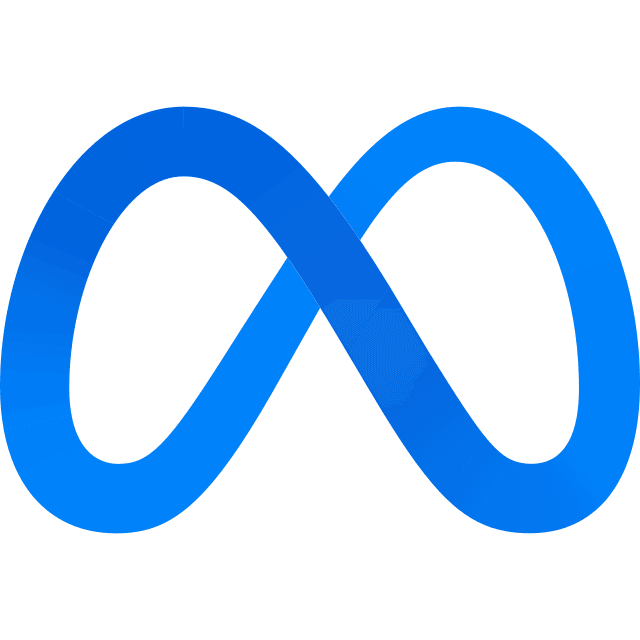}}{#1}}
\newcommand{\LlavaModel}[1]{\modelmark{\includegraphics[height=0.34cm]{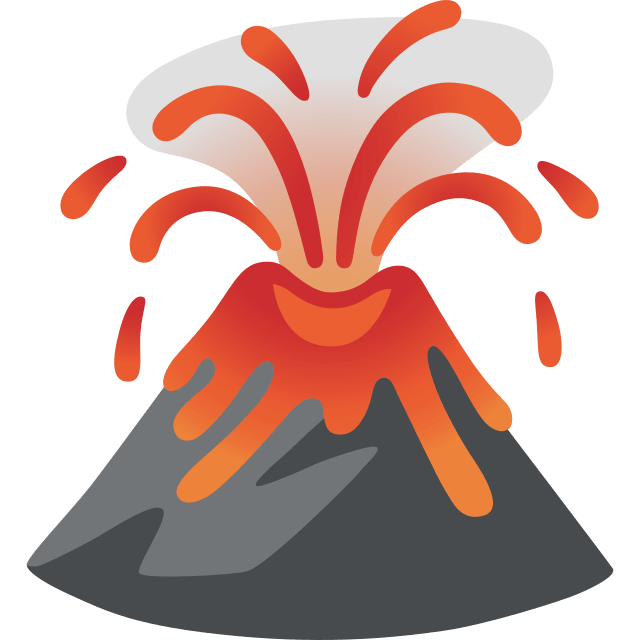}}{#1}}
\newcommand{\GPTModel}[1]{\modelmark{\includegraphics[height=0.34cm]{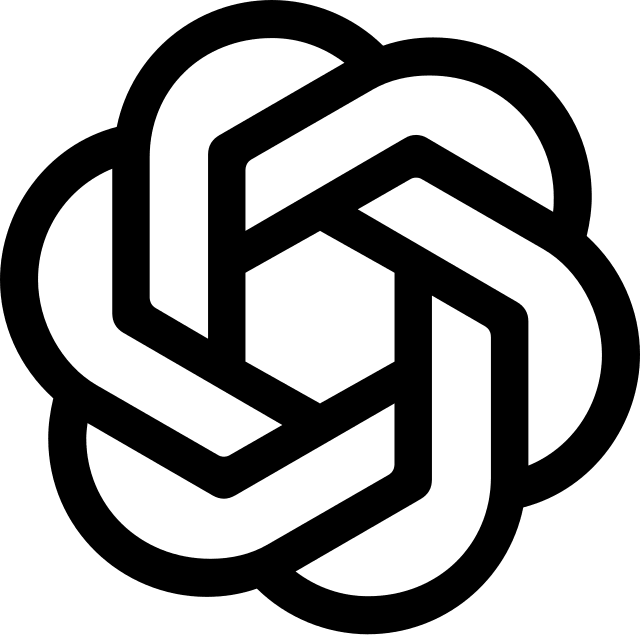}}{#1}}
\newcommand{\GeminiModel}[1]{\modelmark{\includegraphics[height=0.34cm]{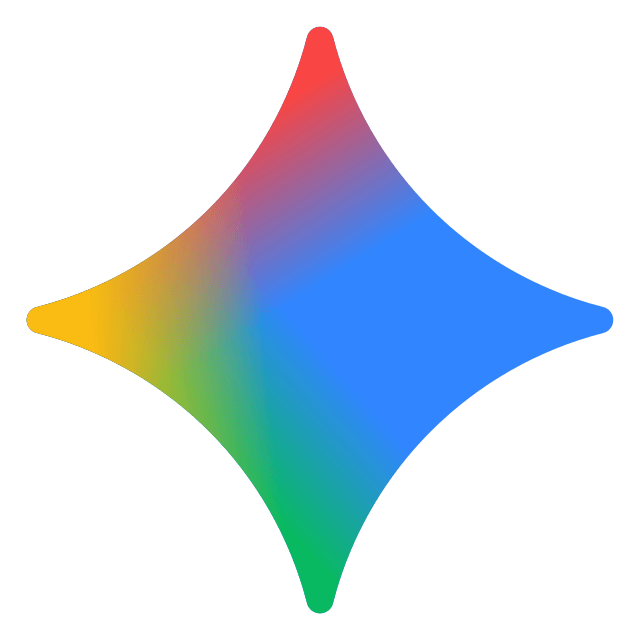}}{#1}}

\tcbset{promptbox/.style={enhanced,boxrule=0.55pt,arc=1.2mm,left=1.4mm,right=1.4mm,top=1.1mm,bottom=1.1mm,fonttitle=\bfseries\footnotesize,coltitle=white,colback=black!2}}

\begin{document}

\title{Front-to-Back: Benchmarking Vision-Language Models for Asymmetric Cross-View Vehicle Re-Identification}
\titlerunning{Low-Resource Asymmetric Vehicle Re-ID}

\author{
Moseli Mots'oehli\inst{1,2} \and
Thulani Babeli\inst{1}
}

\authorrunning{M. Mots'oehli and T. Babeli}
\institute{
MindForge AI, Johannesburg, South Africa\\
\and
University of Hawai'i at M\={a}noa, Honolulu, HI, USA
}

\maketitle

\begin{abstract}
Matching the same vehicle across front and rear cameras is difficult because the cameras do not share a view and the vehicle’s appearance changes substantially. We introduce Front2Back-ReID, a benchmark of 500 manually verified vehicle handovers from 20 recording sequences in South Africa. Each example asks a model to match a vehicle highlighted in a front-camera image to the same vehicle among at least three candidates in a later rear-camera image. We evaluate seven zero-shot vision-language models, four image-retrieval baselines, and 25 human participants. Models are tested using full front RGB images, cropped target vehicles, and binary silhouettes. The strongest VLM achieved 76.6\% Rank-1 accuracy on target crops, compared with 74.0\% for the frozen SigLIP2 baseline; this difference was not statistically clear. Human participants achieved 94.0\% accuracy with full images and 92.2\% with target crops. Under our evaluation setup, enabling reasoning improved accuracy across all three input conditions for every model evaluated in both modes. We also found that VLMs generally performed worse on full scenes than on target crops. These results show that general-purpose VLMs do not yet consistently outperform strong visual retrieval for front-to-rear vehicle matching, while humans remain substantially more reliable. \url{https://mind-forge-ai.github.io/Front2Back-ReID/}.
\end{abstract}

\begin{figure*}[htb]
\centering
\includegraphics[width=0.96\textwidth]{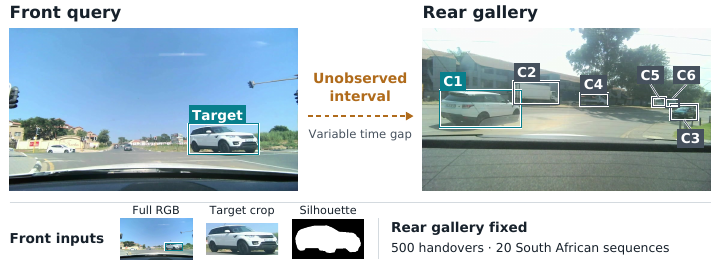}
\caption{A highlighted front-view vehicle disappears through a blind interval and reappears in a closed rear gallery. The task is to select its unique match under full-RGB, crop, or silhouette front evidence.
}
\label{fig:front2back_overview}
\end{figure*}

\section{Introduction}
\label{sec:introduction}

Although the Waymo Open Dataset's End-to-End Driving collection and similar datasets provide images from eight cameras covering the full 360$^\circ$ around the vehicle~\cite{waymo2025endtoend}, resource-constrained settings motivate methods for simpler rigs with fewer views, making identity association across gaps in coverage particularly important. Recent work studies vehicle matching across non-overlapping cameras~\cite{lin2025roundabouthd} and uses temporal features and motion prediction to maintain tracks through occlusions~\cite{pang2023standing}. We focus on front-to-rear matching: identifying a vehicle seen ahead when it later appears behind the ego vehicle, without intervening observations, a fixed inter-view time interval, or a reliable front-to-rear transform.

Such handovers arise during overtakes, pass-bys, turns, merges, and intersection crossings. Large viewpoint changes~\cite{chu2019viewpoint} and new distractors in the rear scene complicate identity matching. We evaluate zero-shot vision-language models (VLMs) on a focused task: \textbf{given one highlighted front target and a closed rear gallery, which candidate is the same vehicle?} Each gallery contains at least three candidates, including exactly one match. This evaluates identity association across an observation gap.

Motivated by diagnostic evaluations of VLM visual grounding~\cite{guan2024hallusionbench}, we examine whether zero-shot VLMs offer an advantage over frozen visual retrieval and how their performance changes with the supplied evidence. We keep the rear gallery unchanged and show the front target in three forms: the full RGB scene, a target crop, and a binary silhouette. These comparisons reveal how models respond to different presentations of the same target. Because context, target scale, and representation also change, they do not isolate the effects of context or reasoning alone.

\dataset{} contains 500 manually verified handovers from 20 South African recording sequences. We evaluate seven VLMs across the three input conditions, four crop-based retrieval controls, and 25 human participants on full-RGB and target-crop trials. The strongest completed VLM achieves 76.6\% Rank-1 accuracy on crops, versus 74.0\% for frozen SigLIP2, but this difference is not statistically significant. Humans achieve 94.0\% on full scenes and 92.2\% on crops. Under the shared prompt protocol, crops outperform full scenes in every informative, VLM comparison.

Our contributions are:
\begin{enumerate}
\item \textbf{A front-to-rear association benchmark:} 500 asymmetric, closed-gallery handovers from 20 South African recording sequences, with manually verified identities and difficulty annotations.
\item \textbf{A diagnostic study of visual evidence:} seven zero-shot VLMs evaluated with full scenes, target crops, and binary silhouettes, alongside four crop-based retrieval controls and a 25-participant human reference. The best completed VLM shows no statistically significant advantage over SigLIP2 on crops, and crops outperform full scenes in every informative, completed VLM comparison.
\end{enumerate}

\section{Related Works}
\label{sec:related_work}

\subsection{Cross-view Vehicle Re-identification}
\label{subsec:Cross-view_vehicle_re-id}

Vehicle Re-ID has largely been studied as retrieval across fixed
surveillance cameras. VehicleID and VERI-Wild match vehicle appearance
across large city camera networks~\cite{liu2016deep,lou2019veriwild}, and
VeRi-776 and CityFlow add spatial and temporal cues to narrow the
search~\cite{liu2016veri,tang2019cityflow}. In these settings, cameras are
static and often elevated, galleries are large, and the recordings come
mostly from Asia, North America, and Europe, with little to no coverage of
Africa. 

Large viewpoint change, especially between front and rear views, is a
well-known difficulty. Methods address it with viewpoint-dependent
metrics~\cite{chu2019viewpoint}, local features and reranking~\cite{wang2020localview}, generated-view adaptation~\cite{wang2021viewpointadaptation}, alignment of visible
regions~\cite{meng2020pven}, view-invariant pretraining~\cite{wang2025vehiclemae}, and fusion of complementary
views~\cite{zheng2023multiquery}. We do not propose another such method.
Instead, we ask how well existing systems use identity cues across views
when evaluated under shared conditions.

\dataset{} changes both the viewpoint setting and the geography. The
cameras ride on the ego vehicle, so a target leaves the front camera's view
and reappears in the rear camera from a very different viewpoint after a
blind interval, among $K_i\geq3$ candidates supplied by surrounding
traffic. Fixed galleries and controlled front evidence let us compare
zero-shot VLMs, frozen retrieval models, and humans on the same decisions.
The recordings come from South African roads, where the mix of vehicles, the layout of the road, and the driving conditions differ from those of existing benchmarks. To our knowledge, \dataset{} is the first vehicle Re-ID benchmark recorded on Southern African roads.

\subsection{Foundation Models for Visual Retrieval}
\label{subsec:Foundation_models_visual_retrieval}

Foundation models are increasingly used for vehicle Re-ID, but usually
after task-specific training; CLIP-ReID, for example, fine-tunes CLIP on
standard person and vehicle benchmarks~\cite{li2023clipreid}. Used frozen,
encoders such as DINOv2~\cite{oquab2024dinov2} and
SigLIP2~\cite{tschannen2025siglip2} are strong general-purpose matchers,
but they are trained to group similar content, not to separate individual
instances: two white sedans may look alike to them. It has been shown that this blind spot carries over to multimodal models, whose errors trace to visually distinct images with similar CLIP embeddings~\cite{tong2024eyes}. Since modern VLMs commonly rely on pretrained visual encoders, we use frozen DINOv2 and SigLIP2 as representation baselines. ranking rear candidates by crop-level embedding similarity therefore lets us ask whether language supervision improves visual identity matching, and whether generative multimodal models add gains beyond embedding similarity on the cross-view vehicle Re-ID task.

\subsection{Visual and Spatial Reasoning in VLMs}
\label{subsec:Visual_spatial_reasoning}

Recent VLMs have gained the abilities our task appears to need.
LLaVA-OneVision is trained across single-image, multi-image, and video
tasks~\cite{li2024onevision}; Qwen2.5-VL processes images at dynamic
resolution and can localize objects~\cite{bai2025qwen25vl}; SpatialVLM
improves spatial reasoning through dedicated
supervision~\cite{chen2024spatialvlm}; and V* uses language-guided search
to find small details in crowded, high-resolution
images~\cite{wu2024vstar}. Applying such models zero-shot to Re-ID in
driving is only beginning: the closest study has a VLM describe each
object crop and matches the descriptions in language
space~\cite{zeroshotsemreid2026}. We instead ask VLMs to compare images directly, with no video or cross-rig geometry to fall back on. Full scenes versus target crops test whether context helps, and running each model with and without reasoning(for applicable models) tests whether extended reasoning improves performance.

\subsection{Diagnostic Benchmarks for Visual Grounding}
\label{subsec:Diagnostic_benchmarks}

Several benchmarks test whether VLMs ground their answers in the image.
HallusionBench uses controlled question pairs to expose hallucination and
visual illusion~\cite{guan2024hallusionbench}, MuirBench pairs multi-image
questions with unanswerable variants~\cite{wang2025muirbench}, and MIHBench
tests whether models keep object identity consistent across images,
finding that errors depend on how many images are shown and where
distractors appear~\cite{li2025mihbench}. These benchmarks probe grounding
by varying the question or the image set. \dataset{} instead holds the
rear gallery fixed and varies only the form of the front evidence (RGB,
crop, or silhouette), on real handovers where viewpoint reverses and
traffic supplies the distractors. Frozen retrieval baselines and human evaluators on the same galleries show whether an instance is solvable from appearance alone.

\section{Benchmark and Task}
\label{sec:dataset}
\begin{figure*}[htb]
    \centering
    \includegraphics[width=\textwidth]{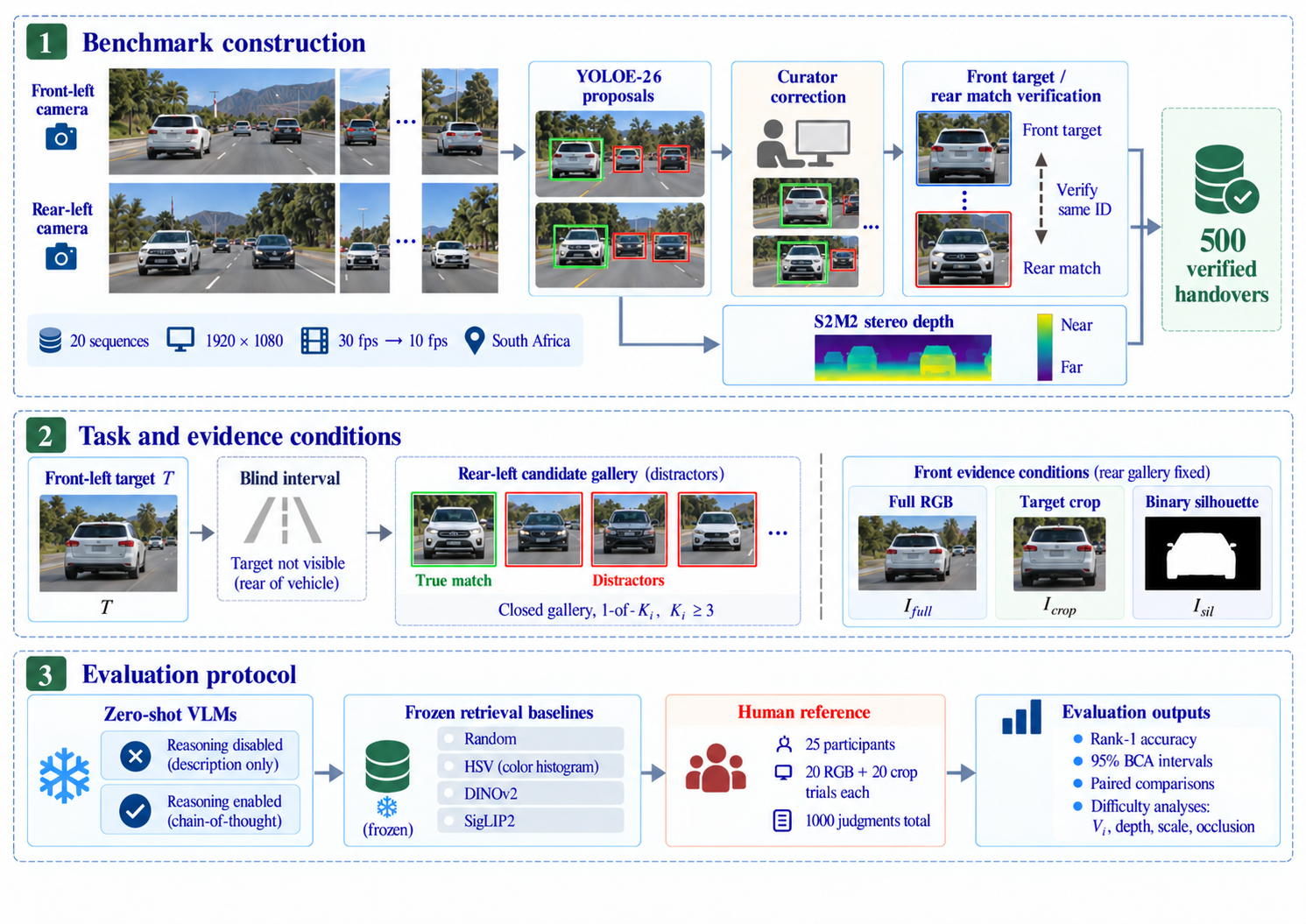}
    \caption{{Overview of \dataset{}. Curators correct and verify YOLOE-26 proposals from front and rear recordings, yielding 500 handovers
    (S\textsuperscript{2}M\textsuperscript{2} depth aids difficulty analysis only). A matcher must find the unique rear-gallery match of a highlighted front-left target after a blind interval, given full RGB, crop, or silhouette. We compare zero-shot VLMs, frozen retrieval baselines, and humans on the same decisions.}}
    \label{fig:overview}
\end{figure*}

\subsection{Task Definition}
\label{subsec:task_def}
An overview of Front2Back-ReID is shown in Fig.~\ref{fig:overview}, including benchmark construction, task setup, evidence conditions, and evaluation. Each example pairs a front-left image containing one highlighted vehicle with a later rear-left image containing $K_i\geq3$ labelled candidates. The task is to select the unique candidate matching the front target; all others are distractors (Fig.~\ref{fig:front2back_overview}). No intervening video, front-to-rear transform, or range estimates are supplied to matchers.

\subsection{Data Collection and Annotation}
\label{sec:annotation}
\begin{figure}[thb]
    \centering
    \includegraphics[width=0.27\linewidth]{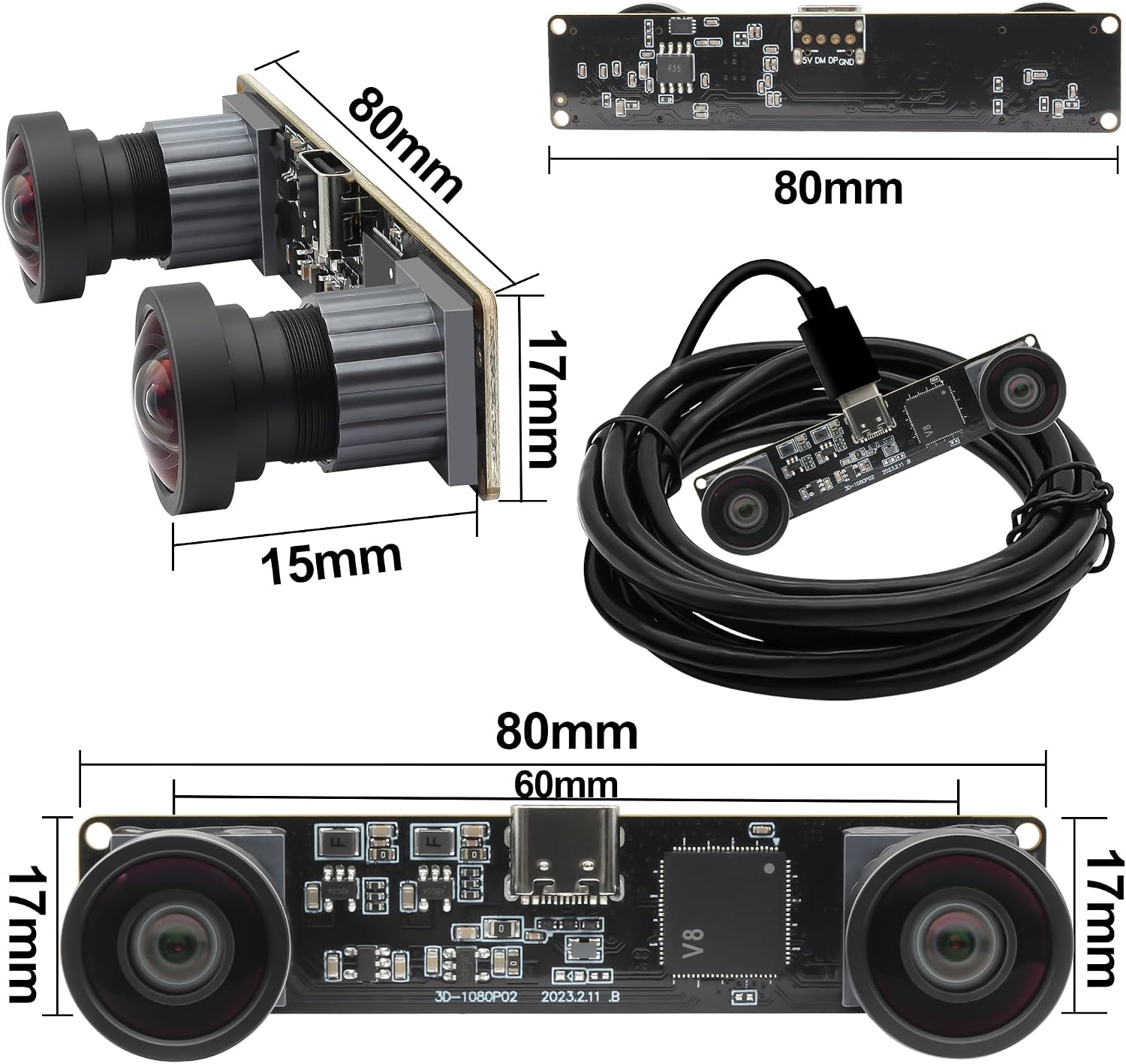}
    \caption{Standard stereo cameras used for data collection.}
    \label{fig:mmlove_camera}
\end{figure}

We recorded 20 sequences on different days and at different locations. We mounted two low-cost MMLove stereo cameras, shown in Fig.~\ref{fig:mmlove_camera}, on the vehicle: one on the windscreen facing forward and one on the rear window facing backward. Each costs under US\$100, has a 60\,mm stereo baseline, and records at $1920\times1080$ pixels and 30\,fps. Camera streams were software-synchronized during recording and downsampled to 10\,fps for processing. 

Curators corrected initial YOLOE-26~\cite{wang2025yoloe} generated bboxes and Silhouette mask polylines, selected front targets, and verified their rear-view matches. We estimated the range of the front-target from the rectified front stereo using the variant S of S\textsuperscript{2}M\textsuperscript{2}~\cite{min2025s2m2} and the valid median disparity within each target bbox. These distance estimates are used to support difficulty analysis only.

\subsection{Construction and statistics}
\label{sec:dataset_characteristics}

\begin{table}[htb]
\centering
\small
\caption{\dataset{} summary. Each record is one closed-gallery handover.}
\label{tab:dataset_summary}
\setlength{\tabcolsep}{4pt}
\renewcommand{\arraystretch}{1.08}
\begin{tabularx}{\linewidth}{@{}>{\raggedright\arraybackslash}X>{\raggedleft\arraybackslash}p{0.40\linewidth}@{}}
\toprule
\textbf{Property} & \textbf{Value} \\
\midrule
Hand-labeled associations & 500 \\
Recordings  & 20 \\
Task & Closed gallery, 1-of-$K_i, K_i\ge 3$,  \\
Canonical views & Front-left, rear-left \\
Model conditions & 3: RGB, crop, mask \\
Human reference conditions & 2: RGB, crop \\
Human reference judgments & $25\times40=1000$ \\
\bottomrule
\end{tabularx}
\end{table}

\begin{figure}[htb]
\centering
\begin{minipage}{0.245\linewidth}
\centering
\includegraphics[width=\linewidth]{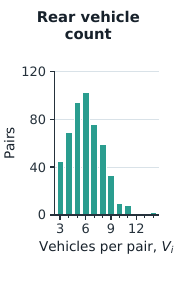}
\end{minipage}\hfill
\begin{minipage}{0.245\linewidth}
\centering
\includegraphics[width=\linewidth]{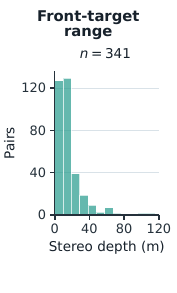}
\end{minipage}\hfill
\begin{minipage}{0.245\linewidth}
\centering
\includegraphics[width=\linewidth]{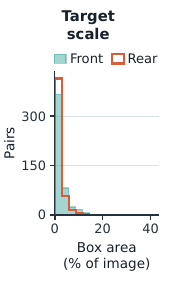}
\end{minipage}\hfill
\begin{minipage}{0.245\linewidth}
\centering
\includegraphics[width=\linewidth]{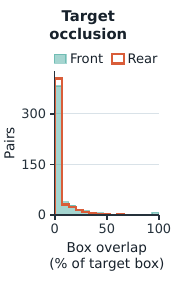}
\end{minipage}
\caption{Benchmark difficulty distributions: vehicle-only candidate count, approximate front-target range, normalized target scale, and occlusion. Range uses the 341 pairs with valid S\textsuperscript{2}M\textsuperscript{2} stereo depth; the other distributions use all 500 pairs.}

\label{fig:dataset_distributions}
\end{figure}

Beyond the setup in Table~\ref{tab:dataset_summary}, difficulty comes from
both sides of a handover: how many rear vehicles compete with the target,
and how far, small, or occluded the target is
(Fig.~\ref{fig:dataset_distributions}). We measure rear-side competition
with $V_i$, the number of vehicles among the rear candidates. This differs
from the gallery size $K_i$, the number of candidates a matcher chooses
from, on 164 pairs, so the 3, 4--5, and $\geq6$ groups hold 45/163/292
pairs under $V_i$ but 38/127/335 under $K_i$. All count-stratified results
use $V_i$. We further detail this distinction, scene composition, and recording durations in Supplementary Secs.~S1 and S8.

\subsection{Difficulty Stratification}
\label{sec:difficulty_stratification}

We group pairs by what makes matching hard: the number of rear vehicles $V_i$, and the target's size, occlusion, and truncation in each view. Size split into three equal beans, and front-target depth into five equal
bins over the 341 pairs with valid S\textsuperscript{2}M\textsuperscript{2}
estimates. These difficulty attributes are used only for analysis and are not given as input to models or evaluators. We provide the thresholds in Supplementary Sec.~S1 and analyze depth and
size sensitivity in Sec.~S7.

\section{Evaluation Protocol}
\label{sec:evaluation_protocol}

\begin{figure}[htb]
\centering
\includegraphics[width=1.01\linewidth]{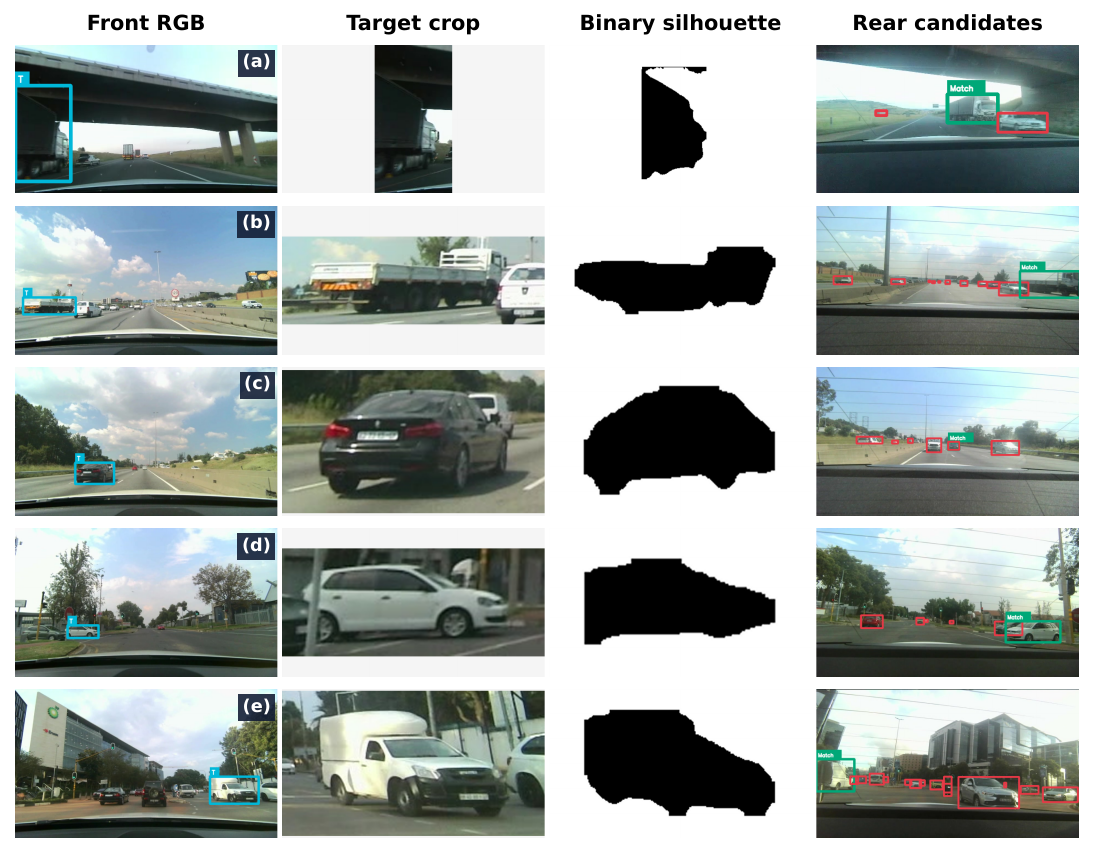}
\caption{Representative benchmark inputs. Columns show the marked full-RGB query, the RGB target crop, the binary silhouette used in the mask condition, and the rear candidate gallery. The silhouette was supplied as black foreground on white background and contained no color, texture, lights, windows, or markings. Green match boxes and red distractor boxes are reader annotations in this figure only.}
\label{fig:qualitative_inputs}
\end{figure}

\begin{figure*}[t]
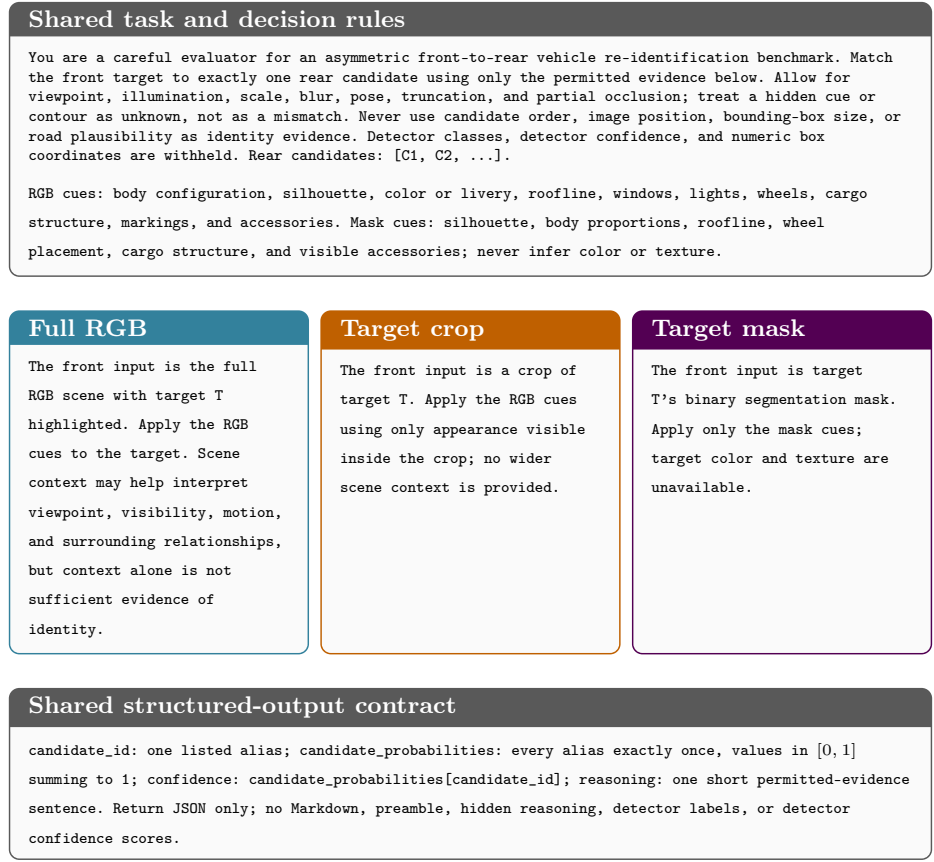

\centering
\begin{tcolorbox}[promptbox,title=Shared task and decision rules,colframe=black!65,colbacktitle=black!65,width=\linewidth]
{\ttfamily\fontsize{6.5}{7.4}\selectfont\raggedright
You are a careful evaluator for an asymmetric front-to-rear vehicle re-identification benchmark. Match the front target to exactly one rear candidate using only the permitted evidence below. Allow for viewpoint, illumination, scale, blur, pose, truncation, and partial occlusion; treat a hidden cue or contour as unknown, not as a mismatch. Never use candidate order, image position, bounding-box size, or road plausibility as identity evidence. Detector classes, detector confidence, and numeric box coordinates are withheld. Rear candidates: [C1, C2, \ldots].\par\smallskip
\textbf{RGB cues:} body configuration, silhouette, color or livery, roofline, windows, lights, wheels, cargo structure, markings, and accessories. \textbf{Mask cues:} silhouette, body proportions, roofline, wheel placement, cargo structure, and visible accessories; never infer color or texture.}
\end{tcolorbox}
\vspace{0.5mm}
\begin{tcbraster}[raster columns=3, raster equal height, raster column skip=1.5mm]
\begin{tcolorbox}[promptbox,title=Full RGB,colframe=cyan!55!black,colbacktitle=cyan!55!black]
{\ttfamily\fontsize{6.5}{7.4}\selectfont\raggedright
The front input is the full RGB scene with target T highlighted. Apply the RGB cues to the target. Scene context may help interpret viewpoint, visibility, motion, and surrounding relationships, but context alone is not sufficient evidence of identity.}
\end{tcolorbox}
\begin{tcolorbox}[promptbox,title=Target crop,colframe=orange!75!black,colbacktitle=orange!75!black]
{\ttfamily\fontsize{6.5}{7.4}\selectfont\raggedright
The front input is a crop of target T. Apply the RGB cues using only appearance visible inside the crop; no wider scene context is provided.}
\end{tcolorbox}
\begin{tcolorbox}[promptbox,title=Target mask,colframe=violet!65!black,colbacktitle=violet!65!black]
{\ttfamily\fontsize{6.5}{7.4}\selectfont\raggedright
The front input is target T's binary segmentation mask. Apply only the mask cues; target color and texture are unavailable.}
\end{tcolorbox}
\end{tcbraster}
\vspace{0.5mm}
\begin{tcolorbox}[promptbox,title=Shared structured-output contract,colframe=black!65,colbacktitle=black!65,width=\linewidth]
{\ttfamily\fontsize{6.5}{7.4}\selectfont candidate\_id: one listed alias; candidate\_probabilities: every alias exactly once, values in $[0,1]$ summing to 1; confidence: candidate\_probabilities[candidate\_id]; reasoning: one short permitted-evidence sentence. Return JSON only; no Markdown, preamble, hidden reasoning, detector labels, or detector confidence scores.}
\end{tcolorbox}
\caption{Prompt specification used for all VLM evaluations. The upper and lower boxes give the decision rules and output schema common to all conditions; the middle boxes state the visual evidence available in each condition.}
\label{fig:model_prompts}
\end{figure*}

Each model sees the same rear gallery with three versions of the front
evidence, shown in Fig.~\ref{fig:qualitative_inputs}: the full RGB scene
with the target marked, an RGB crop of the target, and a binary silhouette.
The crop and silhouette each remove something, but not only that. The crop
drops the surrounding scene but also enlarges the target, and the
silhouette keeps only shape, dropping color and texture. We therefore treat
differences between conditions as effects of how the input is presented,
not of a single cue. Rear candidates are labeled only by their aliases
(C1, C2, \ldots); box coordinates, detector classes and confidences, and other
diagnostic metadata are neither drawn on the images nor included in the
prompts. Figure~\ref{fig:model_prompts} gives the full prompt. 
All conditions share the same task rules and output format; only the description of the front evidence changes. We set temperature to 0 where the model allows it.

\section{Models, Baselines, and Human Reference}
\label{sec:models}

\paragraph{Controls.}
Four non-generative baselines isolate progressively richer visual cues.
Uniform sampling over the $K_i$ candidates fixes chance. An HSV histogram
matched by Bhattacharyya distance~\cite{swain1991color,hafner1995efficient}
measures what color alone recovers. DINOv2 ViT-B/14~\cite{oquab2024dinov2}
and SigLIP2 Base Patch16-224~\cite{tschannen2025siglip2} rank candidate
crops by cosine similarity of frozen embeddings, measuring how much
identity is recoverable without language. 
\paragraph{VLMs.}
Table~\ref{tab:model_roster} lists the evaluated VLMs, chosen to span the options most accessible to resource-constrained researchers and
practitioners: a small open model run locally,
larger open models served through a hosted API, and closed frontier models current as of June 2026 as an upper
reference.

\paragraph{Human reference.}
We recruited twenty-five adults, each of whom completed 40 trials (20 RGB,
20 crop), selecting the matching rear bbox within a 25-minute limit and
without AI assistance. We assign pairs to participants with a seeded random procedure, constrained so that each participant's trials are balanced across difficulty levels, scene conditions, and source videos. Every pair receives one
judgment per condition from different participants; no participant sees
both versions of a pair. All sessions were completed, yielding 1000
judgments. 

\begin{table}[htb]
\centering
\scriptsize
\caption{Evaluated VLMs. Experiments were run between \textbf{24 June and 5 July 2026} using the listed models. Parameter counts unavailable from providers are marked n/d.}
\label{tab:model_roster}
\setlength{\tabcolsep}{3.5pt}
\renewcommand{\arraystretch}{1.08}
\begin{tabularx}{\linewidth}{@{}>{\raggedright\arraybackslash}Xlll>{\raggedleft\arraybackslash}p{0.12\linewidth}@{}}
\toprule
\textbf{Model} & \textbf{Family} & \textbf{Access} & \textbf{Provider} & \textbf{Params} \\
\midrule
\multicolumn{5}{@{}l}{\textit{Open-family models}} \\
\LlavaModel{LLaVA-OneVision 0.5B} & Open   & Local & Local            & 0.5B \\
\LlamaModel{Llama 4 Scout}        & Open   & API   & Groq             & 17B/109B \\
\QwenModel{Qwen 3.6 27B}         & Open   & API   & Groq             & 27B \\
\addlinespace[3pt]
\multicolumn{5}{@{}l}{\textit{Closed hosted}} \\
\GPTModel{GPT-5.5}              & Closed & API   & OpenAI           & n/d \\
\GPTModel{GPT-5.4 mini}         & Closed & API   & OpenAI           & n/d \\
\GeminiModel{Gemini 2.5 Pro}       & Closed & API   & Google AI Studio & n/d \\
\GeminiModel{Gemini 2.5 Flash}     & Closed & API   & Google AI Studio & n/d \\
\bottomrule
\end{tabularx}
\end{table}

\section{Evaluation Metrics}
\label{sec:metrics}

For batch $i$, the target appears exactly once among $K_i$ candidates. The primary metric is Rank-1 accuracy:
\begin{equation}
  \mathrm{Rank}\text{-}1 =
  \frac{1}{N}
  \sum_{i=1}^{N}
  \mathbb{1}[y_i = \arg\max_k p_{i,k}].
\end{equation}
Chance is $1/K_i$. Because galleries are event-local, full-dataset retrieval rank is not meaningful.

We report two-sided 95\% bias-corrected and accelerated (BCa) confidence
intervals~\cite{efron1987bca} from 100{,}000 bootstrap resamples over pairs
(seed 32025). Model Outputs that could not be parsed or named an
unlisted candidate count as incorrect. Human accuracy and decision time are reported per condition, each bootstrapped over its 500 judgments; human--model comparisons are paired
over shared pairs. Since pairs from the same recording or participant may be correlated, so these intervals may be somewhat narrow.

\section{Results and Analysis}
\label{sec:results_analysis}

Tables~\ref{tab:nonreasoning_results} and~\ref{tab:reasoning_results} report reasoning-disabled and reasoning-enabled VLM results, respectively. Bracketed values denote 95\% BCa confidence intervals. All reported model results are based on complete $N=500$ runs. Some Qwen reasoning-enabled runs remained incomplete after multiple retries because of repeated inference failures, including request errors, timeouts, and invalid or missing outputs. We therefore exclude these runs from the results and document them separately as Supplementary material, Sec.~S9. The human-reference row is based on the completed 25-participant study and 40 question pairs per person.

We organize our analysis around three questions: how retrieval and reasoning affect VLM performance, how humans compare in accuracy and response time, and which visual factors drive difficulty for humans and models.

\begin{table*}[tbh]
\centering
\scriptsize
\caption{Rank-1 accuracy (\%) without reasoning, with 95\% BCa intervals. Retrieval controls use crops only. Gemini 2.5 Pro is excluded here because its reasoning cannot be disabled. \textbf{Bold} and \underline{underline} mark the best and second-best
non-human result per column.
$\Delta_{\mathrm{ctx}}$ is full RGB minus target crop: every VLM except the C1-only LLaVA-OneVision loses
accuracy when given the full scene, while humans gain slightly. On crops,
the best VLMs perform on par with the frozen SigLIP2 encoder, and all
remain far below humans.}
\label{tab:nonreasoning_results}
\setlength{\tabcolsep}{5pt}
\renewcommand{\arraystretch}{1.14}
\begin{tabularx}{\textwidth}{@{}>{\raggedright\arraybackslash}X*{3}{>{\centering\arraybackslash}p{0.17\textwidth}}>{\centering\arraybackslash}p{0.07\textwidth}@{}}
\toprule
\textbf{Method} & \textbf{Full RGB} & \textbf{Target crop} & \textbf{Silhouette} & $\boldsymbol{\Delta_{\mathrm{ctx}}}$ \\
\midrule
\multicolumn{5}{@{}l}{\textit{Retrieval controls}} \\
Random gallery   & -- & \scoreci{17.8}{14.4}{21.2} & -- & -- \\
HSV histogram    & -- & \scoreci{47.6}{43.0}{51.8} & -- & -- \\
DINOv2 ViT-B/14  & -- & \scoreci{49.4}{44.8}{53.6} & -- & -- \\
SigLIP2 Base     & -- & \scoreci{74.0}{69.8}{77.6} & -- & -- \\
\addlinespace[3pt]
\multicolumn{5}{@{}l}{\textit{Open-family VLMs}} \\
\LlavaModel{LLaVA-OneVision 0.5B} & \scoreci{27.4}{23.4}{31.2} & \scoreci{27.4}{23.4}{31.2} & \scoreci{27.4}{23.4}{31.2} & $0.0$ \\
\LlamaModel{Llama 4 Scout}        & \scoreci{54.8}{50.2}{59.0} & \scoreci{61.4}{56.8}{65.4} & \secondscoreci{34.2}{30.0}{38.2} & $-6.6$ \\
\QwenModel{Qwen 3.6 27B}          & \bestscoreci{63.6}{59.2}{67.6} & \secondscoreci{74.6}{70.4}{78.0} & \scoreci{32.2}{28.0}{36.2} & $-11.0$ \\
\addlinespace[3pt]
\multicolumn{5}{@{}l}{\textit{Closed hosted VLMs}} \\
\GPTModel{GPT-5.5}             & \secondscoreci{58.4}{53.8}{62.4} & \bestscoreci{74.8}{70.6}{78.2} & \bestscoreci{36.8}{32.4}{40.8} & $-16.4$ \\
\GPTModel{GPT-5.4 mini}        & \scoreci{44.6}{40.0}{48.8} & \scoreci{58.4}{53.8}{62.4} & \scoreci{28.6}{24.6}{32.6} & $-13.8$ \\
\GeminiModel{Gemini 2.5 Flash} & \scoreci{50.6}{46.0}{54.8} & \scoreci{64.0}{59.6}{68.0} & \scoreci{25.2}{21.4}{29.0} & $-13.4$ \\
\midrule
\rowcolor{gray!12}
Human reference & \scoreci{94.0}{91.6}{95.8} & \scoreci{92.2}{89.6}{94.4} & -- & $+1.8$ \\
\bottomrule
\end{tabularx}
\end{table*}

\begin{table*}[tbh]
\centering
\scriptsize
\caption{Rank-1 accuracy (\%) with medium-effort reasoning. $\Delta_r$ is the change from the same model without reasoning
(Table~\ref{tab:nonreasoning_results}). Every model evaluated in both modes improves, but
full RGB still trails the crop by 11--14 points. \textbf{Bold} and
\underline{underline} mark the best and second-best model.}
\label{tab:reasoning_results}
\setlength{\tabcolsep}{2.5pt}
\renewcommand{\arraystretch}{1.2}
\begin{tabular*}{\textwidth}{@{\extracolsep{\fill}}l cc cc cc c@{}}
\toprule
  & \multicolumn{2}{c}{\textbf{Full RGB}}
  & \multicolumn{2}{c}{\textbf{Target crop}}
  & \multicolumn{2}{c}{\textbf{Silhouette}}
  & \\
\cmidrule(lr){2-3}\cmidrule(lr){4-5}\cmidrule(lr){6-7}
\textbf{Method} & Rank-1 & $\Delta_r$ & Rank-1 & $\Delta_r$ & Rank-1 & $\Delta_r$ & $\boldsymbol{\Delta_{\mathrm{ctx}}}$ \\
\midrule
\GPTModel{GPT-5.5}             & \bestscoreci{62.8}{58.2}{66.8}   & \rgain{+4.4}  & \bestscoreci{76.6}{72.6}{80.0}   & \rgain{+1.8}  & \bestscoreci{43.8}{39.4}{48.0}   & \rgain{+7.0}  & $-13.8$ \\
\GPTModel{GPT-5.4 mini}        & \scoreci{58.8}{54.2}{62.8}       & \rgain{+14.2} & \scoreci{71.0}{66.6}{74.6}       & \rgain{+12.6} & \secondscoreci{40.0}{35.6}{44.2} & \rgain{+11.4} & $-12.2$ \\
\GeminiModel{Gemini 2.5 Pro}   & \secondscoreci{60.6}{56.0}{64.6} & \rgain{--}    & \secondscoreci{71.8}{67.4}{75.4} & \rgain{--}    & \scoreci{34.6}{30.4}{38.6}       & \rgain{--}    & $-11.2$ \\
\GeminiModel{Gemini 2.5 Flash} & \scoreci{54.4}{49.8}{58.6}       & \rgain{+3.8}  & \scoreci{67.2}{62.8}{71.0}       & \rgain{+3.2}  & \scoreci{32.0}{27.8}{36.0}       & \rgain{+6.8}  & $-12.8$ \\
\midrule
\rowcolor{gray!12}
Human reference & \scoreci{94.0}{91.6}{95.8} & -- & \scoreci{92.2}{89.6}{94.4} & -- & -- & -- & $+1.8$ \\
\bottomrule
\end{tabular*}
\end{table*}

\begin{figure*}[tbh!]
\centering
\includegraphics[width=\textwidth]{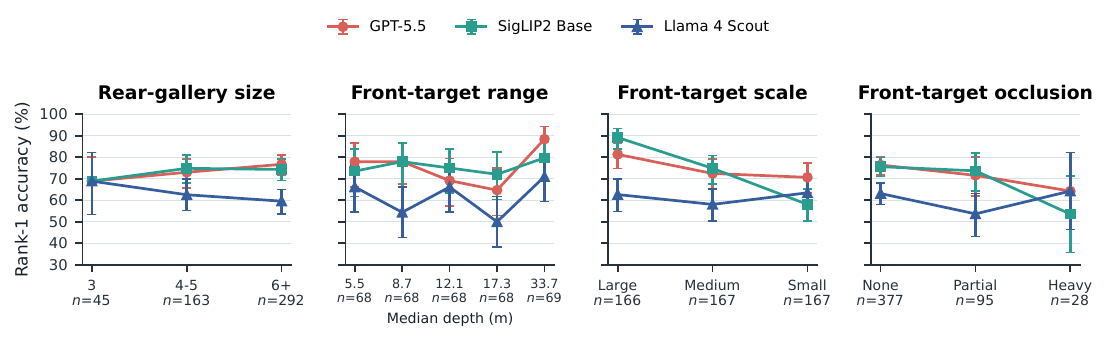}
\caption{Target-crop Rank-1 accuracy without reasoning, by vehicle-only candidate count,
front-target range, scale, and occlusion. The panel labeled ``Rear-gallery size'' uses vehicle-only count $V_i$, not gallery size $K_i$. The range panel uses the 341
pairs with valid S\textsuperscript{2}M\textsuperscript{2} depth, split into
five equal-frequency bins; the others use all 500. Smaller targets and heavier occlusion generally lower accuracy; range and candidate-count trends are weaker.}
\label{fig:difficulty_strata}
\end{figure*}

\begin{figure*}[tbh!]
\centering
\includegraphics[width=\textwidth]{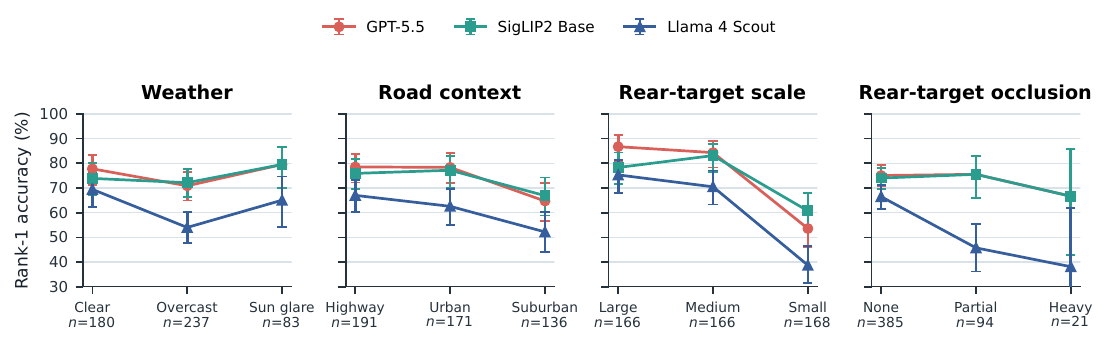}
\caption{Target-crop accuracy without reasoning by weather, road context, and correct rear-target scale and occlusion. GPT-5.5 and Llama~4 Scout fall sharply on small or heavily occluded targets, while SigLIP2 stays flatter. Construction is omitted from the road-context panel (two pairs).}
\label{fig:context_strata}
\end{figure*}

\begin{figure}[tbh!]
\centering
\includegraphics[width=\linewidth]{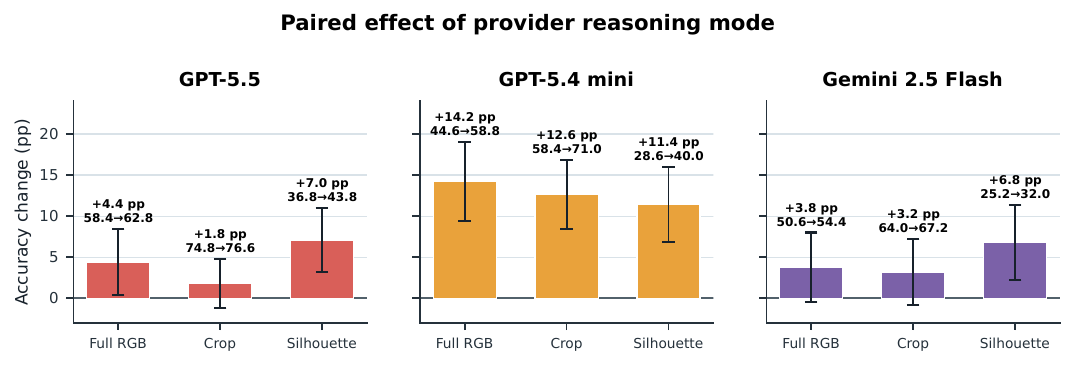}
\caption{Change in Rank-1 accuracy with medium-effort reasoning for complete matched runs. Differences and uncertainty intervals are paired by annotation; intervals crossing zero do not establish a directional gain.}
\label{fig:reasoning_effect}
\end{figure}

\paragraph{Reasoning helps selectively and does not beat frozen retrieval.}
With reasoning enabled, GPT-5.5 is the best crop model, 2.6 points above
SigLIP2. The paired 95\% BCa interval, $[-2.2, 7.4]$, includes zero:
GPT-5.5 alone is correct on 81 pairs and SigLIP2 alone on 68, so the
advantage is not significant. We find that reasoning gains are also uneven (Fig.~\ref{fig:reasoning_effect}): GPT-5.4 mini gains 11--14 points in every condition, whereas GPT-5.5 and
Gemini 2.5 Flash gain most on silhouettes and show minimal gain on crops. We include the list of all paired contrasts in Supplementary Sec.~S4.

The same pattern holds for every VLM except LLaVA-OneVision: without
reasoning, crops beat full scenes by 6.6--16.4 points, and silhouettes
fall 27.2--42.4 points below crops. Because each change also alters target
scale or representation, these gaps reflect how the input is presented,
not context alone. LLaVA-OneVision selects C1 on every trial, so its
identical 27.4\% scores simply equal the share of pairs whose match is C1,
even though its outputs are valid JSON. We detail model explanations and a pair missed by all ten complete crop
runs in Supplementary Sec.~S6, and document the incomplete Qwen runs in
Sec.~S9.

\begin{figure}[tbh]
\centering
\includegraphics[width=0.75\linewidth]{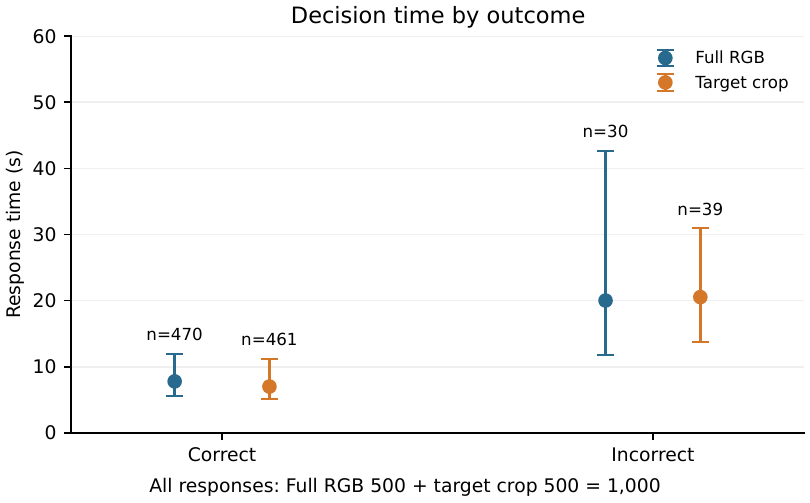}
\caption{Human decision time by outcome. Points show medians and bars the interquartile range (IQR). All responses contribute, including those above 60\,s.}
\label{fig:human_response_time}
\end{figure}

\begin{figure*}[tbh!]
\centering
\includegraphics[width=\textwidth]{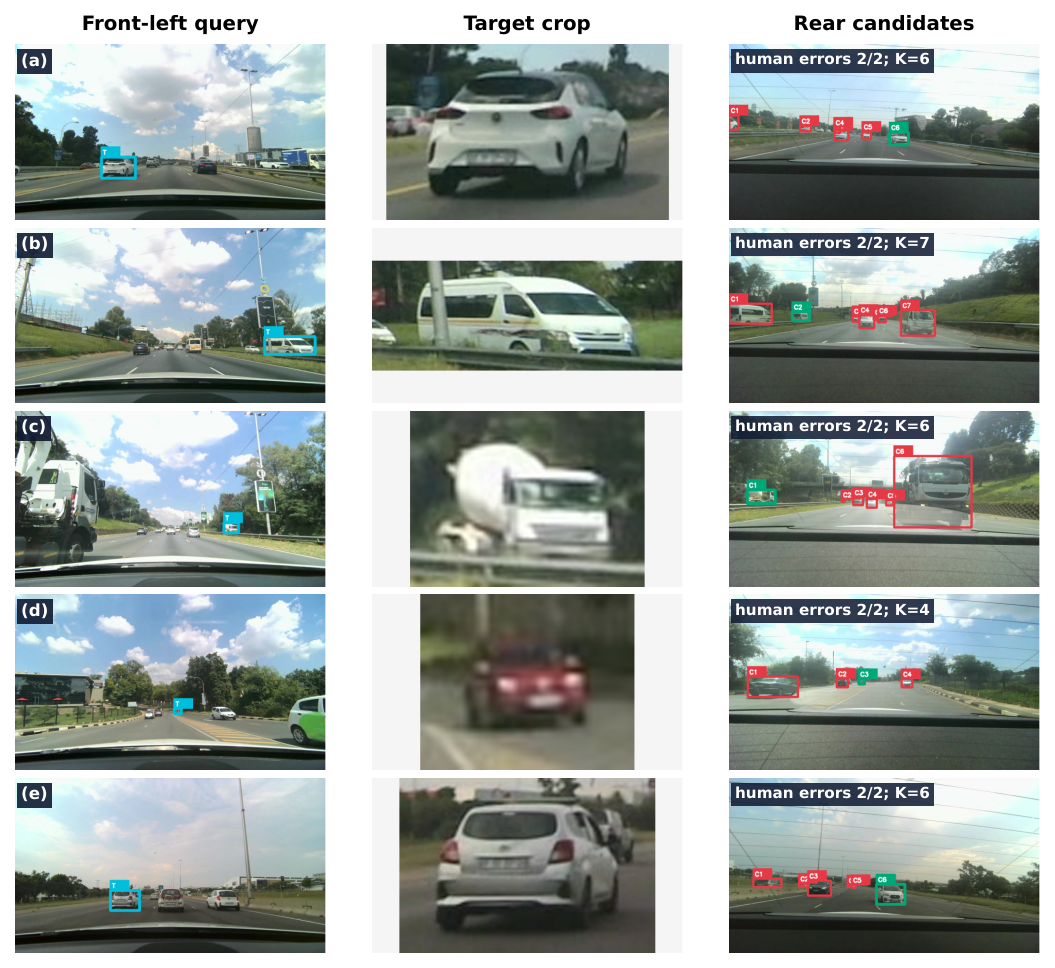}
\caption{Five of the ten pairs missed in both human conditions. Columns show the full query, target crop, and rear gallery; green marks the match and red the distractors. Each pair received one judgment per condition.}
\label{fig:human_hard}
\end{figure*}

\paragraph{Humans remain far ahead.}
Humans reach 94.0\% on full RGB and 92.2\% on crops, 31.2 and 15.6 points
above reasoning-enabled GPT-5.5, and 18.2 points above SigLIP2 on crops.
Incorrect answers also take longer (Fig.~\ref{fig:human_response_time}):
median decision time is 20.0 versus 7.8\,s for incorrect and correct
answers on full RGB, and 20.5 versus 7.0\,s on crops. This may reflect harder pairs or hesitation. Fig.~\ref{fig:human_hard} shows five of the ten pairs missed in both conditions; with only two judgments per pair, this is a coarse difficulty signal. Supplementary Sec.~S5 repeats the timing analysis without responses over 60\,s that may be due to our tool's interface issues.

\begin{figure}[tbh!]
\centering
\includegraphics[width=\linewidth]{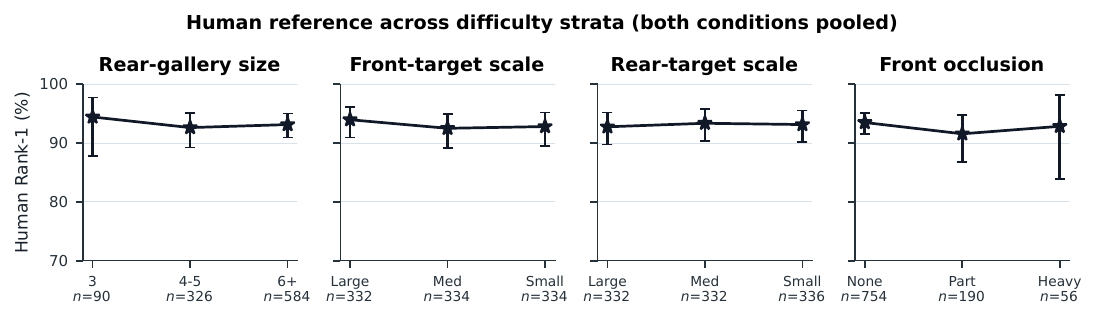}
\caption{Human Rank-1 accuracy by difficulty stratum, pooled over full RGB and target crops (1{,}000 judgments from 25 participants). Uncertainty intervals resample responses. Accuracy remains high even for small or heavily occluded targets, where models degrade (Figs.~\ref{fig:difficulty_strata} and~\ref{fig:context_strata}).}
\label{fig:human_strata}
\end{figure}

\paragraph{Small rear targets separate humans from models.}
Target scale separates humans from models more than vehicle count does
(Figs.~\ref{fig:difficulty_strata}, \ref{fig:context_strata},
and~\ref{fig:human_strata}). On the smallest third of rear targets, human
accuracy (pooled over both conditions) stays at 93.2\%, while crop
accuracy without reasoning drops for every model and falls to 38.7\% for
Llama 4 Scout. Human accuracy is nearly flat across these strata, suggesting that models
miss identity cues people can still use, although humans also miss some
pairs in both conditions (Fig.~\ref{fig:human_hard}). Front-target depth shows no consistent trend across the five depth bins;
we analyze depth sensitivity further in Supplementary Sec.~S7.

\section{Limitations}
\label{sec:limitations}

\dataset{} covers 500 handovers from 20 recordings in five South African
cities, about 1.9 hours of driving. While limited, it is designed for controlled
comparison of methods on real handovers, not for broad claims about other
countries, fleets, or sensors. Daytime scenes dominate (97.6\%), so night
and adverse weather are underrepresented, and depth is available for 341
pairs, which limits range analysis to that subset. Our evidence conditions
change more than one factor at a time; a crop, for example, removes
context but also enlarges the target, so we interpret them as effects of
input presentation. Finally, our intervals treat pairs as independent, and
each pair has one human judgment per condition. Resampling by recording
and collecting more judgments per pair would give more conservative
intervals and a sharper human reference.

\section{Conclusion}
\label{sec:conclusion}

We introduced \dataset{}, a benchmark of 500 verified front-to-rear
vehicle handovers recorded with low-cost cameras on South African roads,
where no calibrated transform links the front and rear views. Holding each
rear gallery fixed, we varied the front evidence between full scene,
target crop, and silhouette, and compared zero-shot VLMs, frozen retrieval
models, and 25 human participants on the same decisions. Three findings
stand out. 

First, the best VLM, reasoning-enabled GPT-5.5, reaches 76.6\%
on crops, not significantly above the frozen SigLIP2 encoder at 74.0\%,
while humans reach 92.2--94.0\%. Secondly, full scenes yield lower accuracy than 
target crops for every informative VLM run by 6.6--16.4 points without reasoning, whereas
humans score 1.8 points higher with full RGB scenes. Cropping also changes
target scale, so this difference cannot be attributed to context alone;
reasoning does not remove the observed VLM gap. 

Third, small rear targets separate models from humans, whose
accuracy stays nearly flat across difficulty strata. \dataset{} thus
offers a focused test of cross-view identity association for settings
where calibrated multi-sensor rigs are out of reach. We plan to
extend this work to night driving, diverse weather conditions and more regions, adding the intervening video so methods can use motion between views, and developing methods that make better use of scene context for identity matching, rather than being distracted by it.

\section*{Acknowledgements}
We thank the 25 participants in the human evaluation for their time, and Dalitso Chomey for valuable discussions on the paper.

\clearpage
\bibliographystyle{splncs04}
\bibliography{main}

\clearpage
\definecolor{reasonblue}{RGB}{30,82,120}
\newcommand{\reasontext}[1]{\textcolor{reasonblue}{\emph{#1}}}
\newcolumntype{Y}{>{\raggedright\arraybackslash}X}

\setcounter{section}{0}
\setcounter{figure}{0}
\setcounter{table}{0}
\setcounter{equation}{0}
\renewcommand{\thesection}{S\arabic{section}}
\renewcommand{\thefigure}{S\arabic{figure}}
\renewcommand{\thetable}{S\arabic{table}}
\renewcommand{\theequation}{S\arabic{equation}}
\renewcommand{\theHsection}{supp.\arabic{section}}
\renewcommand{\theHfigure}{supp.\arabic{figure}}
\renewcommand{\theHtable}{supp.\arabic{table}}
\renewcommand{\theHequation}{supp.\arabic{equation}}

\begin{center}
{\Large\bfseries Supplementary Material}
\end{center}

We provide additional dataset and protocol details, numerical comparisons, and examples of model and human errors. Unless stated otherwise, accuracy intervals use 100{,}000 BCa resamples (seed 32025). Model intervals resample pairs; pooled human-stratum intervals resample individual responses. Latency bars show interquartile ranges (IQRs).

\section{Dataset composition and difficulty definitions}
\label{sup:composition}
The 500 handovers come from 20 South African recordings. Gallery size $K_i$ counts every candidate shown to evaluators, including non-vehicle road users. The difficulty analyses instead use $V_i$, the number of vehicles among those candidates after class and ego-vehicle filtering. The counts differ on 164 pairs. For $K_i$, the 3, 4--5, and $\geq6$ groups contain 38, 127, and 335 pairs; for $V_i$, they contain 45, 163, and 292. Candidate aliases and ground-truth matches are unchanged, and detector classes are hidden from evaluators. Count-stratified plots use $V_i$ and cannot, by themselves, establish how accuracy varies with gallery size $K_i$.

\begin{table}[htb]
\centering
\small
\caption{Vehicle-only candidate counts and visibility (500 pairs). $V_i$ differs from actual gallery cardinality.}
\label{tab:dataset_difficulty_stats}
\setlength{\tabcolsep}{7pt}
\renewcommand{\arraystretch}{1.06}
\begin{tabularx}{\linewidth}{@{}>{\raggedright\arraybackslash}X@{\hspace{10pt}}S[table-format=3.0]@{\hspace{16pt}}S[table-format=2.1]@{}}
\toprule
\textbf{Attribute} & {\textbf{Count}} & {\textbf{Share (\%)}} \\
\midrule
\multicolumn{3}{@{}l}{\textit{Vehicle-only candidate count}} \\
\quad $V_i=3$ & 45 & 9.0 \\
\quad $V_i=4$--$5$ & 163 & 32.6 \\
\quad $V_i\geq6$ & 292 & 58.4 \\
\addlinespace[3pt]
\multicolumn{3}{@{}l}{\textit{Front-target visibility}} \\
\quad No occlusion & 377 & 75.4 \\
\quad Partial occlusion & 95 & 19.0 \\
\quad Heavy occlusion & 28 & 5.6 \\
\quad Boundary truncation & 66 & 13.2 \\
\addlinespace[3pt]
\multicolumn{3}{@{}l}{\textit{Rear-target visibility}} \\
\quad No occlusion & 385 & 77.0 \\
\quad Partial occlusion & 94 & 18.8 \\
\quad Heavy occlusion & 21 & 4.2 \\
\quad Boundary truncation & 48 & 9.6 \\
\bottomrule
\end{tabularx}
\end{table}

\begin{table}[htb]
\centering
\small
\caption{Scene-context composition across the 500 handovers.}
\label{tab:scene_context_stats}
\setlength{\tabcolsep}{6pt}
\renewcommand{\arraystretch}{1.04}
\begin{tabularx}{\linewidth}{@{}>{\raggedright\arraybackslash}p{0.24\linewidth}>{\raggedright\arraybackslash}X@{\hspace{8pt}}S[table-format=3.0]@{\hspace{14pt}}S[table-format=2.1]@{}}
\toprule
\textbf{Factor} & \textbf{Category} & {\textbf{Count}} & {\textbf{Share (\%)}} \\
\midrule
Time of day & Day & 488 & 97.6 \\
& Dawn or dusk & 10 & 2.0 \\
& Night & 2 & 0.4 \\
\addlinespace[3pt]
Weather & Overcast & 237 & 47.4 \\
& Clear & 180 & 36.0 \\
& Sun glare & 83 & 16.6 \\
\addlinespace[3pt]
Road context & Highway & 191 & 38.2 \\
& Urban & 171 & 34.2 \\
& Suburban & 136 & 27.2 \\
& Construction & 2 & 0.4 \\
\bottomrule
\end{tabularx}
\end{table}

\paragraph{Scale and range bins.}
Normalized target area is divided into tertiles at 0.93\% and 2.26\% in the front view (bin counts 166/167/167) and 0.66\% and 1.57\% in the rear view (166/166/168). Occlusion is labeled none, partial, or heavy; truncation denotes contact with an image boundary. The 341 pairs with valid approximate front-stereo depth form five near-equal bins (68/68/68/68/69), separated at 6.91, 9.91, 14.54, and 22.28\,m. Depth plots exclude pairs without valid stereo estimates; they do not substitute inverse image area for depth. These annotations support trial balancing and difficulty analysis but are not supplied to matchers.

\section{Additional protocol details and difficult cases}
\label{sup:protocol}
\begin{figure}[htb]
\centering
\includegraphics[width=\linewidth]{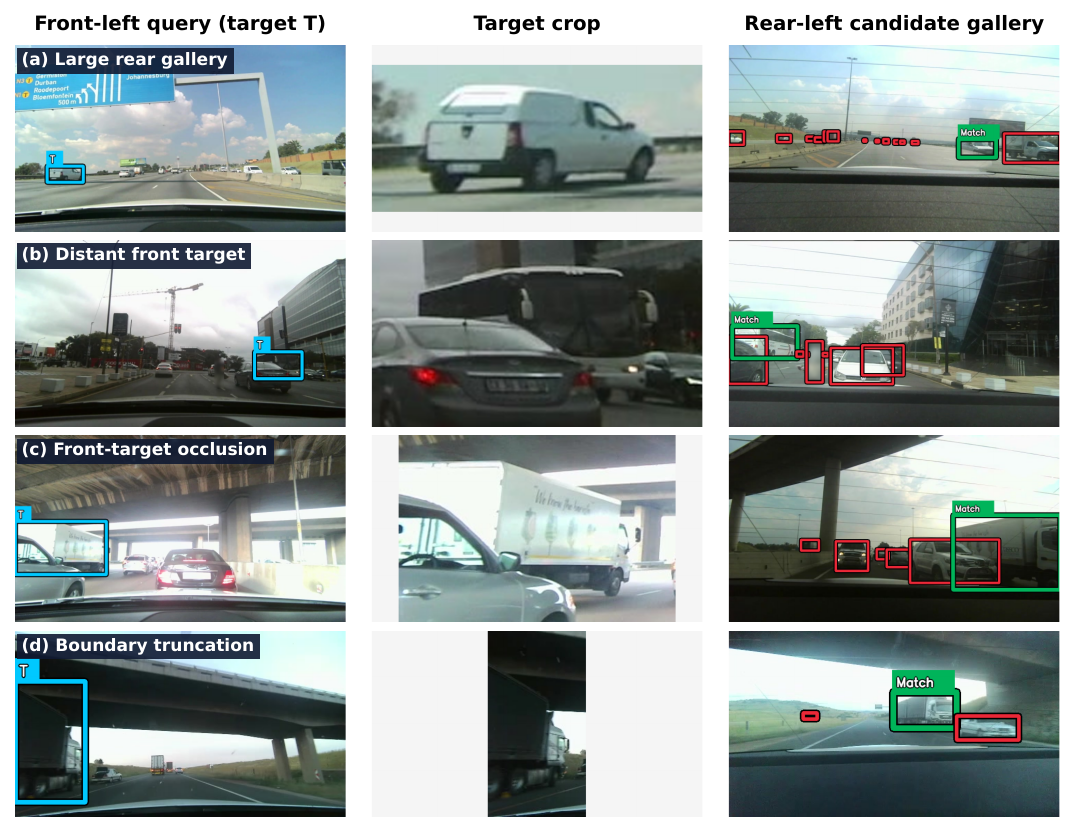}
\caption{Difficult handovers from \dataset{}. Columns show the marked full-RGB query, target crop, and fixed rear gallery. Human participants saw one of the two front inputs and did not see the answer annotations. Green marks the match and red the distractors for the reader; detector classes are omitted and people are blurred. Rows show (a)~many vehicle candidates ($V_i{=}14$), (b)~a distant front target ($\approx113$\,m approximate stereo depth), (c)~front-target occlusion ($\approx61\%$), and (d)~boundary truncation.}
\label{fig:difficulty_plate}
\end{figure}

Each front-evidence condition uses the same rear gallery. Candidate aliases remain visible, while coordinates, detector classes, and detector confidence scores are withheld. The prompt instructs models not to use candidate order, position, box size, or road plausibility as identity evidence, although visual ordering remains apparent. A crop removes scene context and changes the target's effective scale. Figure~\ref{fig:difficulty_plate} illustrates four challenging cases under this protocol.

\section{Evidence-condition comparisons}
\label{sup:ablation}
From the reasoning-disabled results in Table~\ref{tab:nonreasoning_results} of the main paper, crops improve Rank-1 accuracy over full RGB by 6.6 points for Llama 4 Scout, 11.0 for Qwen 3.6 27B, 16.4 for GPT-5.5, 13.8 for GPT-5.4 mini, and 13.4 for Gemini 2.5 Flash. Their respective losses from crop to silhouette are 27.2, 42.4, 38.0, 29.8, and 38.8 points. These are observed differences, without separate significance tests. They reflect changes in input presentation, including target scale and available cues. LLaVA-OneVision returned C1 throughout, so its unchanged accuracy does not show insensitivity to the missing evidence.

\section{Paired reasoning comparisons}
\label{sup:reasoning}

Figure~\ref{fig:reasoning_effect} in the main paper compares medium-effort reasoning with reasoning disabled on the same pairs. Its 95\% BCa intervals use 100{,}000 paired resamples. \GPTModel{GPT-5.4 mini} gained 14.2 points on full RGB $[9.4, 19.0]$, 12.6 on crops $[8.4, 16.8]$, and 11.4 on silhouettes $[6.8, 16.0]$. \GPTModel{GPT-5.5} gained 4.4 points on full RGB $[0.4, 8.4]$ and 7.0 on silhouettes $[3.2, 11.0]$; \GeminiModel{Gemini 2.5 Flash} gained 6.8 on silhouettes $[2.2, 11.4]$.

The remaining paired intervals cross zero: GPT-5.5 crops, $+1.8$ points $[-1.2, 4.8]$; Gemini 2.5 Flash full RGB, $+3.8$ $[-0.4, 8.0]$; and Gemini 2.5 Flash crops, $+3.2$ $[-0.8, 7.2]$. Thus, reasoning helped GPT-5.4 mini across all three inputs, but the evidence for gains in the other matched cells is mixed. In particular, the best crop result (GPT-5.5) has no clear paired reasoning gain.

\section{Human accuracy, latency, and errors}
\label{sup:human}
Each of the 25 participants answered 20 full-RGB and 20 crop trials, without seeing the same pair twice. Each pair received one judgment per condition from different participants. The main-paper timing plot uses all 1{,}000 judgments: 470 correct and 30 incorrect on full RGB, and 461 correct and 39 incorrect on crops.

These give accuracies of 94.0\% and 92.2\%, respectively. Pooled human-stratum intervals resample individual responses, without clustering by pair or participant; they describe uncertainty under that sampling choice.

\subsection{Decision latency}
\label{sup:latency}

Across all responses, median decision time was 8.09\,s (IQR 5.75--13.08) for full RGB and 7.27\,s (5.29--12.60) for crops. These times include interface use and individual differences; they do not measure reasoning alone.

\begin{figure}[htb]
\centering
\includegraphics[width=\linewidth]{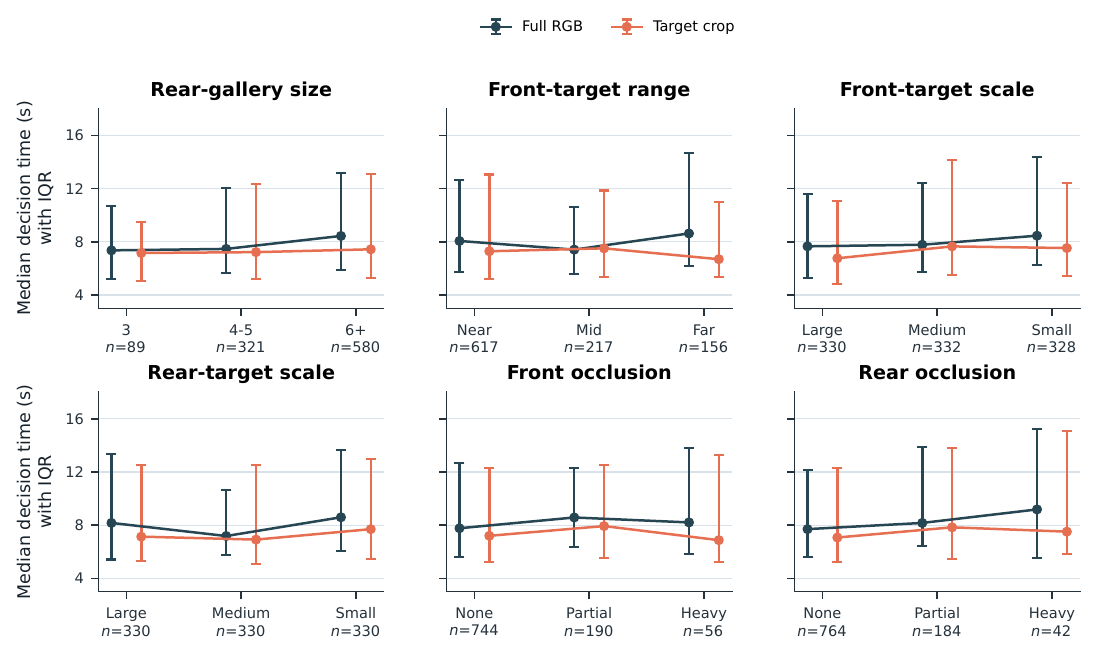}
\caption{Human decision time by difficulty stratum after the 60\,s cutoff, shown separately for full RGB and crops. Points are medians and bars are IQRs; counts pool both input conditions.}
\label{fig:sup_latency_difficulty}
\end{figure}

\begin{figure}[htb]
\centering
\includegraphics[width=\linewidth]{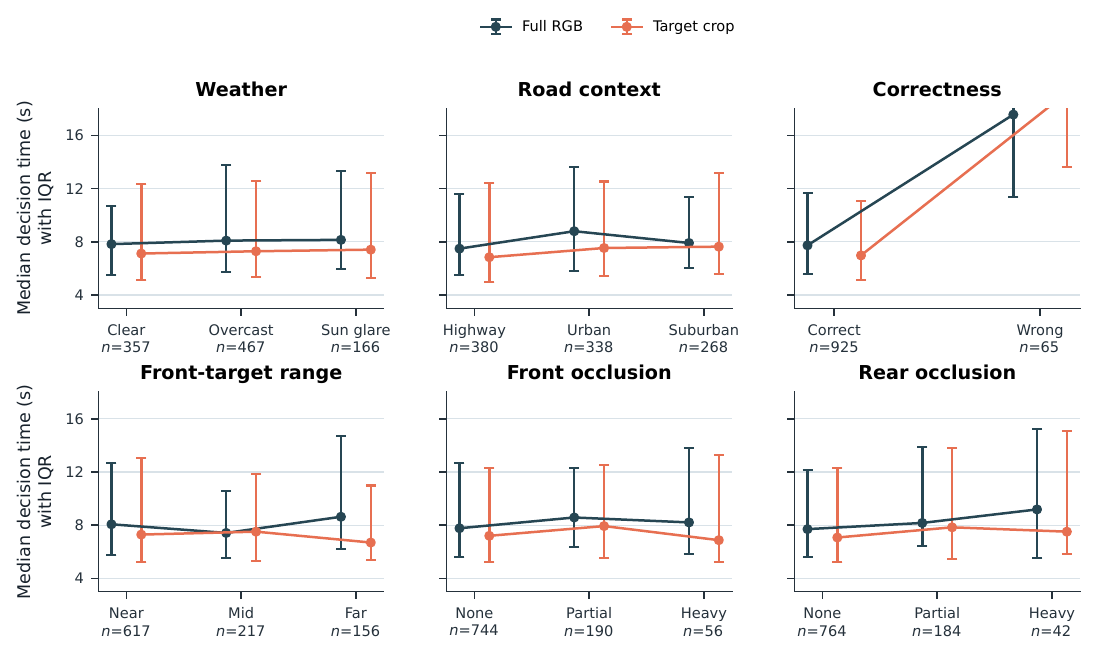}
\caption{Human decision time by scene context, correctness, and visibility after the 60\,s cutoff. Incorrect responses take longer; weather, road context, and occlusion show smaller median differences.}
\label{fig:sup_latency_context}
\end{figure}

The supplementary latency plots exclude 10 responses longer than 60\,s, leaving 492 full-RGB and 498 crop responses. This differs from the main-paper timing plot, which includes all 1{,}000 responses. After filtering, median times for correct and incorrect responses are 7.74 and 17.57\,s on full RGB ($n=465$ and $27$), and 6.98 and 19.21\,s on crops ($n=460$ and $38$). Without filtering, the corresponding medians are 7.78 and 20.01\,s on full RGB, and 7.00 and 20.53\,s on crops. The 60\,s cutoff removes five correct and three incorrect full-RGB responses, and one correct and one incorrect crop response. Figures~\ref{fig:sup_latency_difficulty} and~\ref{fig:sup_latency_context} split the filtered sample by difficulty and scene context. Their candidate-count labels refer to $V_i$, not gallery size $K_i$. These plots are descriptive; they do not adjust for participant or question order.

\begin{figure}[htb]
\centering
\includegraphics[width=0.72\linewidth]{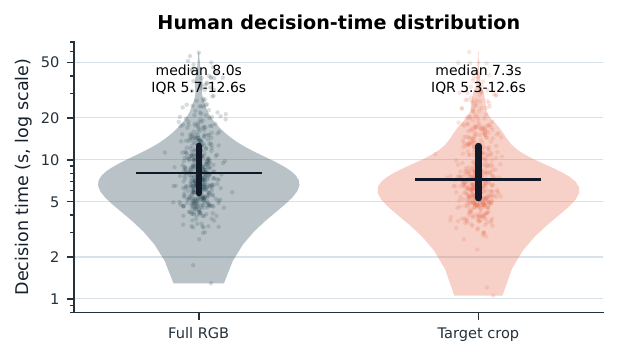}
\caption{Human decision-time distribution by input condition, excluding 10 responses above 60\,s. Violin width shows density; points are responses, horizontal marks are medians, and thick vertical segments are IQRs. Time is shown on a log scale.}
\label{fig:sup_latency_distribution}
\end{figure}

\subsection{Pairs missed in both human conditions}
\label{sup:human_hard}

The main-paper plate shows five of the ten pairs missed in both human conditions. Each pair received one full-RGB and one crop judgment, so these examples show two observed errors per pair. They do not establish majority failure or intrinsic ambiguity.

\section{Qualitative model reasoning examples}
\label{sup:model_examples}

The response schema requested one short sentence citing visual evidence. Table~\ref{tab:sup_reasoning_examples} shows four saved front-crop responses: two correct matches and two errors based on plausible but misleading cues. These sentences are model outputs, not verified explanations of the models' decisions.

\begin{table}[t]
\centering
\caption{Saved model evidence sentences from front-crop trials. ``GT'' is the correct rear-candidate alias; R denotes medium-effort reasoning.}
\label{tab:sup_reasoning_examples}
\footnotesize
\begin{tabularx}{\linewidth}{@{}p{0.20\linewidth}p{0.12\linewidth}p{0.13\linewidth}Y@{}}
\toprule
Model cell & Outcome & Pick / GT & Evidence sentence \\
\midrule
GPT-5.5 (R) & Correct & C1 / C1 & \reasontext{C1 matches the target's white low sedan silhouette with a dark side-window band.} \\
Gemini 2.5 Pro (R) & Correct & C2 / C2 & \reasontext{The target and candidate C2 are both large white tanker trucks.} \\
GPT-5.5 (R) & Wrong & C3 / C1 & \reasontext{C3 matches the target's dark gray SUV body, chrome-accented grille, and headlight shape.} \\
Gemini 2.5 Flash (R) & Wrong & C1 / C2 & \reasontext{The target vehicle and C1 are both white vans with identical yellow and gold stripe livery on their sides.} \\
\bottomrule
\end{tabularx}
\end{table}

\begin{figure}[htb]
\centering
\includegraphics[width=\linewidth]{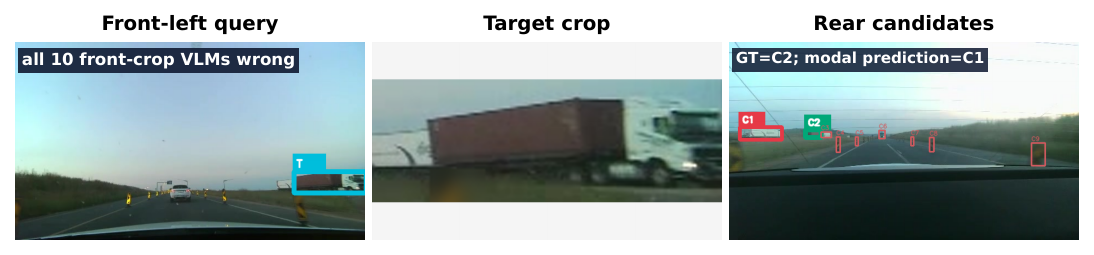}
\caption{Front-crop pair missed by all 10 complete front-crop VLM cells. The correct rear candidate is C2 (green); C1 is the common selected distractor (red). Other candidates are drawn with thin outlines to preserve the small ground-truth box.}
\label{fig:sup_model_failure_pair}
\end{figure}

Figure~\ref{fig:sup_model_failure_pair} and Table~\ref{tab:sup_model_failure_pair} show one pair missed by all ten complete front-crop VLM runs. Every run chose C1, a prominent large truck, while the labeled match was the smaller, more distant C2. LLaVA-OneVision also chose C1, but its run selected C1 on every trial and echoed a prompt template rather than giving visual evidence.

\begin{table}[t]
\centering
\caption{Saved predictions for Fig.~\ref{fig:sup_model_failure_pair}. All runs chose C1; the correct answer is C2. R denotes medium-effort reasoning; confidence is the model's reported value.}
\label{tab:sup_model_failure_pair}
\scriptsize
\begin{tabularx}{\linewidth}{@{}p{0.21\linewidth}ccY@{}}
\toprule
Model cell & Pick & Conf. & Evidence sentence \\
\midrule
GPT-5.5 (R) & C1 & 0.86 & \reasontext{C1 is the only candidate showing a large white-cab tractor with a long reddish-brown box trailer matching the target.} \\
GPT-5.5 (no R) & C1 & 0.72 & \reasontext{C1 shows the same large white tractor with a long reddish-brown box trailer silhouette as the front target.} \\
GPT-5.4 mini (R) & C1 & 0.44 & \reasontext{C1 is the only candidate with a long semi-trailer silhouette and white cab/trailer combination matching the target truck.} \\
GPT-5.4 mini (no R) & C1 & 0.72 & \reasontext{C1 best matches the front target's white boxy vehicle shape with a light cab and long rectangular cargo body.} \\
Gemini 2.5 Pro (R) & C1 & 0.92 & \reasontext{The target is a white tractor pulling a white trailer, and C1 is the only candidate that is a large white truck.} \\
Gemini 2.5 Flash (R) & C1 & 0.92 & \reasontext{The target vehicle is a white semi-truck with a red shipping container, which precisely matches candidate C1.} \\
Gemini 2.5 Flash (no R) & C1 & 1.00 & \reasontext{The target vehicle is a white semi-truck with a red container, which perfectly matches candidate C1.} \\
Qwen 3.6 27B (no R) & C1 & 0.90 & \reasontext{The target vehicle is a white semi-truck cab, which matches the white truck cab visible in candidate C1, whereas all other candidates are small dark vehicles.} \\
Llama 4 Scout (no R) & C1 & 1.00 & \reasontext{The truck's white cab and brown cargo container match.} \\
LLaVA-OneVision 0.5B (no R) & C1 & 0.11 & \reasontext{one short sentence citing visible identity evidence.} \\
\bottomrule
\end{tabularx}
\end{table}
\FloatBarrier

\section{Depth sensitivity}
\label{sup:depth}

\begin{figure}[htb]
\centering
\includegraphics[width=\linewidth]{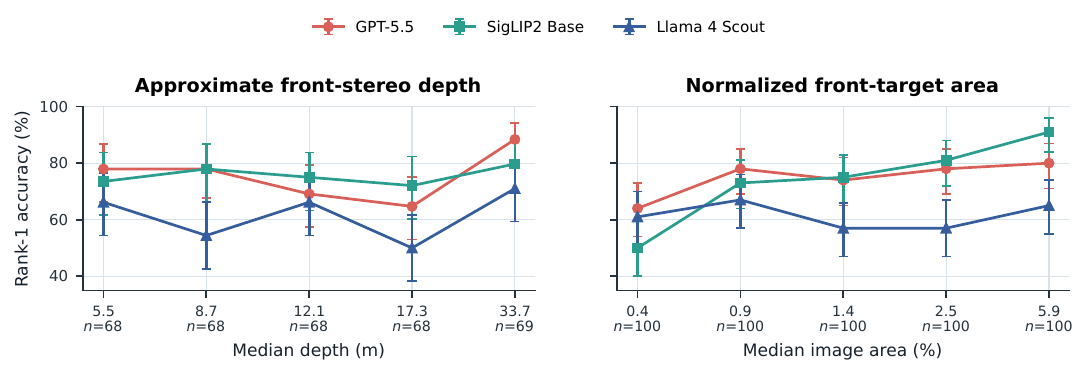}
\caption{Reasoning-disabled crop accuracy in five bins of valid approximate front-stereo depth ($n=341$) and normalized front-target area ($n=500$). Points are placed at each bin's median depth or area. Depth shows no monotonic trend; SigLIP2 improves as target area increases.}
\label{fig:sup_depth}
\end{figure}

Approximate front-stereo depth is available for 341 pairs. Across five equal-frequency depth bins, crop accuracy is non-monotonic for all three plotted methods (Fig.~\ref{fig:sup_depth}, left). Normalized front-target area shows a clearer association with accuracy (right): SigLIP2 rises from 50.0\% in the smallest-area quintile to 91.0\% in the largest, while GPT-5.5 changes less. These binned comparisons support the main paper's scale analysis; they do not isolate scale from other properties of the examples.

\section{Recording-sequence durations}
\label{sup:durations}

\begin{figure}[htb]
\centering
\includegraphics[width=0.62\linewidth]{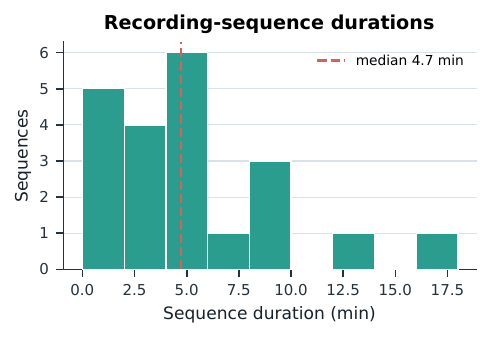}
\caption{Durations of all 21 recorded sequences, including the one with no accepted pair (46 seconds to 17.2 minutes; median 4.7 minutes; approximately 1.9 hours in total).}
\label{fig:sup_durations}
\end{figure}

Twenty-one sequences were recorded, of which 20 contribute accepted pairs to the benchmark. The 21 recordings total approximately 1.9 hours; durations range from 46 seconds to 17.2 minutes, with a median of 4.7 minutes (Fig.~\ref{fig:sup_durations}).

\section{Excluded incomplete runs}
\label{sup:incomplete}

The main-paper interval analyses include only complete $N=500$ runs. The reasoning-enabled \QwenModel{Qwen 3.6 27B} runs reached 470/500 full-RGB, 480/500 crop, and 416/500 silhouette trials before repeated Groq request failures and truncated completions prevented completion. We exclude these incomplete runs. Provider failures and empty or truncated outputs were not scored; non-empty outputs that could not be parsed or selected an unlisted candidate counted as incorrect. This records the limits of these runs under the evaluated settings, not a reliability ranking of model families or providers.

\end{document}